\documentclass[letterpaper, 10 pt, conference]{ieeeconf}  

\IEEEoverridecommandlockouts                              
\usepackage[utf8]{inputenc}
\usepackage[T1]{fontenc}
\usepackage{xcolor}
\usepackage{amsmath}
\usepackage{graphicx,epstopdf}
\usepackage{siunitx}
\usepackage{caption}
\usepackage{subcaption}
\usepackage{svg}
\usepackage{soul}
\usepackage{verbatim}
\usepackage{tikz}
\usepackage{lipsum}

\usepackage[colorinlistoftodos, textsize=small, textwidth=50mm]{todonotes}
\definecolor{lfcolor}{rgb}{.7, 0.1, 0.7}

\newcommand\copyrighttext{%
  \footnotesize \textcopyright2021 IEEE. Personal use of this material is permitted. Permission from IEEE must be obtained for all other uses, in any current or future media, including reprinting/republishing this material for advertising or promotional purposes, creating new collective works, for resale or redistribution to servers or lists, or reuse of any copyrighted component of this work in other works.}
\newcommand\copyrightnotice{%
\begin{tikzpicture}[remember picture,overlay]
\node[anchor=south,yshift=10pt] at (current page.south) {\fbox{\parbox{\dimexpr\textwidth-\fboxsep-\fboxrule\relax}{\copyrighttext}}};
\end{tikzpicture}%
}

\title{\LARGE \bf
Active Safety System for Semi-Autonomous Teleoperated Vehicles
}

\author{Smit Saparia$^{1}$, Andreas Schimpe$^{2}$ and Laura Ferranti$^{3}$ 
\thanks{$^{1}$ S. Saparia was with the Institute for Automotive Technology at the
Technical University of Munich (TUM), Germany and Department of Cognitive Robotics, Delft University of Technology, Delft, The Netherlands
        {\tt\small smit.saparia at outlook.com}}%
\thanks{$^{2}$ A. Schimpe is with the Institute for Automotive Technology at the
Technical University of Munich (TUM), 85748 Garching bei M\"unchen,
Germany. 
        {\tt\small andreas.schimpe at tum.de.}}%
\thanks{$^{3}$ L. Ferranti is with Department of Cognitive Robotics, Delft University of Technology, Delft, The Netherlands
        {\tt\small l.ferranti at tudelft.nl.} Her work is supported by the NWO VENI grant (n. 18165).}%
}

\begin{document}

\maketitle
\thispagestyle{empty}
\pagestyle{empty}

\begin{abstract}

Autonomous cars can reduce road traffic accidents and provide a safer mode of transport. However, key technical challenges, such as safe navigation in complex urban environments, need to be addressed before deploying these vehicles on the market. Teleoperation can help smooth the transition from human operated to fully autonomous vehicles since it still has human in the loop providing the scope of fallback on driver. This paper presents an Active Safety System (ASS) approach for teleoperated driving. The proposed approach helps the operator ensure the safety of the vehicle in complex environments, that is, avoid collisions with static or dynamic obstacles. Our ASS relies on a model predictive control (MPC) formulation to control both the lateral and longitudinal dynamics of the vehicle. By exploiting the ability of the MPC framework to deal with constraints, our ASS restricts the controller's authority to intervene for lateral correction of the human operator's commands, avoiding counter-intuitive driving experience for the human operator. Further, we design a visual feedback to enhance the operator's trust over the ASS. In addition, we propose an MPC's prediction horizon data based novel predictive display to mitigate the effects of large latency in the teleoperation system. We tested the performance of the proposed approach on a high-fidelity vehicle simulator in the presence of dynamic obstacles and latency.

\end{abstract}

\copyrightnotice
\section{INTRODUCTION}
From the introduction of the seat belt to Automatic Emergency Braking (AEB) systems, vehicles have become increasingly safer. Despite the technological improvements, more than 25 thousand fatalities due to road accidents were reported just in the EU in 2017 \cite{EUraodstat}. These reports also highlighted that $\approx$\SI{95}{\percent} of the fatalities were caused by human errors. Fully autonomous vehicles might present the solution to reduce such accidents by removing the human from the loop. Fully autonomous vehicles, however, may still fail in traffic situation a human could easily handle, such as parking places or pedestrian crossing areas. To smooth the transition from human-driven vehicles to fully autonomous vehicles, teleoperated driving can play a fundamental role.

\begin{figure}
    \centering
    \includegraphics[width=\linewidth]{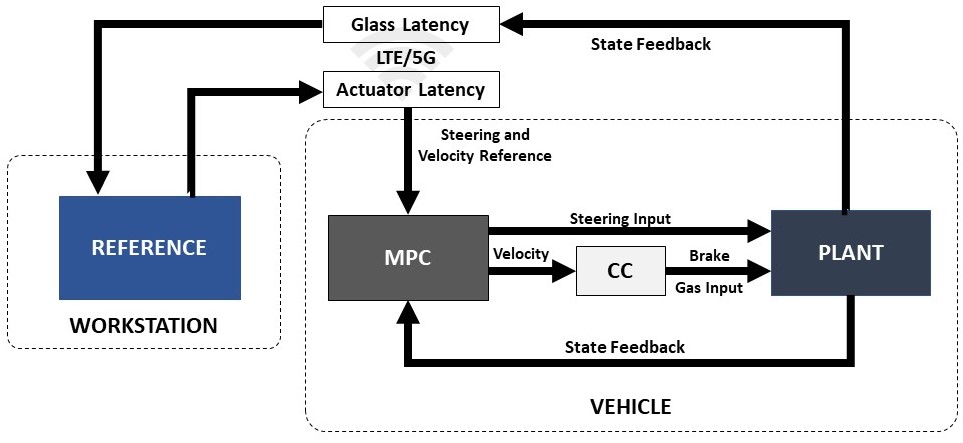}
    \caption{System architecture for Teleoperated Driving showing major components in the closed loop.}
    \label{fig:general architecture}
\end{figure}

In Teleoperated Driving (ToD), a human operator drives the vehicle from a workstation equipped with steering wheel and gas/brake pedals. The generated control commands are transmitted to the vehicle via a mobile network for execution. The operator can view the vehicle's environment on the monitors as captured by vehicle's cameras. This transmission of signals via the mobile network introduces latency. As Fig.~\ref{fig:general architecture} shows, there are two types of latency to consider, that are, \emph{actuator latency} and \emph{glass latency}. The actuator latency is the time taken by the signals to travel from the input devices at workstation to the actuators on the vehicle via the mobile network. The glass latency is the time taken by camera information to travel from camera on the vehicle to the monitors at the workstation via the mobile network. These latency definitions account for the time spent by signals at various stages in hardware as well as in the network. 

ToD has several challenges, such as large latency, complete connection loss, or reduced situation awareness. Large latency in the network can lead to unstable real time control of the vehicle \cite{Pongrac2011GestaltungUE}. To mitigate the effects of latency, techniques like predictive display can be employed. In worst cases, a complete connection loss poses a threat to make teleoperation useless and dangerous at the same time. Nevertheless, strategies like redundant mobile networks are deployed to overcome such an issue \cite{phantom}. Another challenge in ToD is reduced situational awareness. It becomes difficult to judge velocity and location of the obstacles and this problem aggravates in the peripheral region of the view. In tight driving scenarios, the chances of collision increase and hence a driver assistance system to help avoid collision proves beneficial as shown in \cite{andi}. As Fig. \ref{fig:general architecture} shows, the human operator generates steering and velocity commands that are then manipulated by the controller to generate safe collision free control inputs for the vehicle. Since human operator (from the workstation) and the system (with the final authority) control the vehicle, it can be stated as a case of \emph{uncoupled shared control}~\cite{reviewSharedControl}. To mitigate the issue of conflict of control interest between human operator and the system, a feedback is necessary to communicate the intent of the system to the human operator.

\subsection{Contributions}
To address the above challenges, this paper presents an active safety system for collision avoidance using Model Predictive Control (MPC) in a shared control framework specifically for a teleoperated road vehicle, building on the approach presented in \cite{andi}. First, we propose a driver assistance system (Sec. \ref{sec: section2}), based on MPC, for a teleoperated semi-autonomous road vehicle. The primary objective is to track the input references as generated by the human operator and intervene when deemed necessary to avoid collision with static or dynamic obstacles. As compared to \cite{andi}, the approach presented in this paper controls both, the longitudinal and lateral dynamics of the vehicle. In addition, our design introduces a restriction on the controller's authority to override human operator to improve the operator trust on the automation. Furthermore, we introduce potential fields in the MPC formulation to efficiently achieve a collision free manoeuvre. Second, we propose a visual feedback (Sec. \ref{sec: section3}) to communicate system's intent and enhance trust in the human operator over the system to mitigate conflict of control interest. Third, we propose an MPC data-based predictive display technique (Sec. \ref{sec:predictive}) to mitigate effects of high latency in a teleoperation setup. Finally, the proposed approach is validated using a high-fidelity vehicle simulator with the simulated human operator (Sec. \ref{sec: section5}).

\subsection{Related Work}\label{sec:RelatedWork} 
Our work spans across different areas. For this reason, this section provides an overview of the relevant literature on shared control, obstacle avoidance, visual feedback and predictive display with a main focus on MPC-based approaches.

\paragraph*{Shared Control}Several shared control approaches have been proposed in the literature, such as model-based coupled shared control, model-based uncoupled shared control, model-free uncoupled shared control (as defined in~\cite{reviewSharedControl}). In uncoupled shared control, automation has final authority to implement the input to the vehicle unlike in coupled shared control. A model-based shared control models the shared control behaviour, mostly using a driver model. A model-based coupled shared control approach is followed in \cite{Ercan2016TorquebasedSA}, where human driver and steering system models are incorporated into a lateral controller design. Modelling the driver requires actual driving tests as shown in \cite{pred_setinvariance} and the driver model becomes specific to a particular driving style.

A model-free uncoupled shared control approach is followed in ~\cite{anderson}, where a final steering command as a weighted combination of driver and automation input according to a tailored threat metric is computed. Tuning of this threat metric is not trivial and only lateral control was showcased in this work.
Finally, a model-based uncoupled shared control approach is followed in \cite{safeNL}, \cite{sharedcontrol} and \cite{andi}. In \cite{safeNL}, a model predictive contouring control (MPCC) framework is used for automation where center of the lane is considered as a reference path. Another optimization problem considers driver inputs and a decaying function is used to compute a weighted combination of driver and automation inputs. This approach considers both lateral and longitudinal control, but cannot be used for the case of teleoperation where availability of information of lane center via perception module is not reliable. The authors in \cite{sharedcontrol} and \cite{andi} directly incorporate driver's steer commands into the objective function of MPC for only lateral control, where the objective is to match the driver inputs.

\paragraph*{Obstacle Avoidance} Obstacles and ego vehicle can be directly modelled in the constraints of the MPC formulation as \emph{geometrical shapes}, based on which a collision free criterion can be incorporated into the problem formulation of controller, as proposed, for example, in \cite{safeVRU} and \cite{MPClaura} in the context of fully autonomous vehicles. Reactive methods, such as potential fields, can also be incorporated in the MPC formulation instead of explicitly modelling obstacles in the constraints. The authors of~\cite{PF} represented crossable and non-crossable obstacles with different potential fields. The authors of~\cite{andi} modelled the obstacles as higher-order ellipses and formulated a potential function such that the front vehicle corners are pushed away from the obstacles. This potential function is incorporated into the objective of the MPC problem formulation. The advantage of using potential function over spatial constraint is that the size of the optimization problem does not grow with the number of obstacles in the scenario \cite{safeNL,safeVRU}. Therefore, the solution time of the problem is not expected to increase with the increase in the number of obstacles. 

\paragraph*{Visual Feedback} A feedback is necessary for the controller to communicate its intent to the human driver. A haptic feedback on steering wheel is used in \cite{hapticFTM}, to assist human operator in challenging situations and helps prevent collision of teleoperated vehicle. However, haptic feedback on steering wheel can be used only for lateral control. The authors in \cite{VREUGDENHIL2019371} showcased the benefit of having visual feedback along with haptic shared control for obstacle avoidance. Haptic feedback on steering wheel cannot communicate future vehicle state that might affect current driver inputs. This is supposedly compensated by a visual feedback. Monitors in teleoperation can be used to illustrate future vehicle positions. This advantage is exploited in this work.

\paragraph*{Predictive Display} As presented in \cite{ftmlatency}, predictive display can be used to predict ego vehicle position, to compensate for the delay in signal transmission.
This involves projecting the predicted vehicle position images on the video stream. However, these techniques assume constant vehicle velocity and road wheel angle in the predictions.

\section{SHARED CONTROL APPROACH}\label{sec: section2}

\subsection{Prediction Model}
The teleoperated vehicle is expected to be driven at slow speeds with moderate driving. Hence, the prediction model chosen in this approach for the MPC is the following kinematic bicycle model~\cite{Borrelli_kinematic}:
\begin{subequations}\label{eq:kinematic}
\begin{align}
    \dot{x} &= v\cos{(\psi + \beta)}    \label{eq:kinematic1}\\ 
    \dot{y} &= v\sin{(\psi + \beta)}    \label{eq:kinematic2}\\  
    \dot{\psi} &= \frac{v}{l_r}\sin{(\beta)}    \label{eq:kinematic3}\\ 
    \dot{v} &= a    \label{eq:kinematic4}\\
    \beta &= \tan^{-1}{\bigg(\frac{l_r}{l_f + l_r} \tan(\delta)\bigg)}    \label{eq:kinematic4}
\end{align}
\end{subequations}
where $x$ and $y$ are coordinates of the center of mass (CoM) of the vehicle, $\psi$ is the heading angle of the vehicle in the inertial frame, $v$ is the vehicle velocity, $l_f$ and $l_r$ are the distances between CoM and front and rear axles respectively. $\beta$ is the angle that the velocity vector makes with the longitudinal axis of the vehicle, $a$ is the acceleration of the CoM and $\delta$ is the road wheel angle (RWA) of the front wheels. To simplify the notation, we define the following states and input quantities, that are, $z := [x,y,\psi, \delta, v]$, and  $u := [\dot{\delta}, a]$ respectively.

\subsection{Modelling Ego Vehicle and Rectangular Obstacles}
The collision avoidance criteria are incorporated in the MPC problem formulation by approximating the ego vehicle and the rectangular obstacle as geometric shapes. The ego vehicle is represented with four longitudinally shifted circles of radius $r$, along its longitudinal axis ~\cite{MPClaura}. The benefit of representing the ego vehicle with circles instead of just using front vehicle corners as in \cite{andi} for collision avoidance is that it ensures that the ego vehicle stays collision free from the obstacles with a downside of an increase in the computational burden. A rectangular obstacle is represented with a higher order ellipse. The implicit function, as proposed in \cite{andi}, that describes the shape of the obstacle in inertial frame is given by,
\begin{equation}\label{eq:ellipse}
\resizebox{0.9\hsize}{!}{$
    e_{i}^m(x_{c_{i}},y_{c_{i}}) = \bigg(\frac{R(\phi)(x_{c_{i}} - x_{\textrm{obs}}^m(t))}{\alpha_{\textrm{maj}}}\bigg)^n + \bigg(\frac{R(\phi)(y_{c_{i}} - y_{\textrm{obs}}^m(t))}{\beta_{\textrm{min}}}\bigg)^n - 1
    $}
\end{equation}
for $i$ $\in$ $\{1,2,3,4\}$. Here, $(x_{c_{i}}, y_{c_{i}})$ represents the center of the $i^{th}$ ego vehicle circle and $(x_{\textrm{obs}}^m(t), y_{\textrm{obs}}^m(t))$ represents the center of the $m^{th}$ obstacle. $R(\phi)$ is the rotation matrix, depending on the heading $\phi$ of the obstacle. The semi-axis of the $\alpha$ and $\beta$ and the even order $n$ of the ellipse help define the size of the bounding ellipse. Therefore, we define 
\begin{equation}
    \alpha_{\textrm{maj}} = a + r \quad \text{and} \quad \beta_{\textrm{min}} = b + r, 
\end{equation}
where $a := f \cdot L/2,$ $b = f \cdot B/2$, $f = \sqrt[n]{2}$ and $r$ is the radius of the ego vehicle. $L$ and $B$ are the length and breadth of the obstacle respectively.

\subsection{Potential Field for Collision Avoidance}\label{subsec: PF}
Building on \cite{andi}, we rely on a repulsive potential field approach for collision avoidance. The potential field $P$ is a sum of individual potential functions for $m$ obstacles for all the $i$ circles of the ego vehicle, that is,
\begin{equation}
\label{eq:potential_function_all}
    P = \sum_{m} \sum_{i} P_{i}^m (x,y),
\end{equation}
where the potential function, for the $i^{th}$ circle for the $m^{th}$ obstacle is as follows
\begin{gather}\label{eq:pot_function}
    P_{i}^m (x,y) = \frac{\tau}{(e_{i}^m(x_{c_{i}},y_{c_{i}})+1)^{\rho}}.
\end{gather}
In the equation above, $\tau$ and $\rho$ are the design parameters that specifies the strength and slope of the potential function. It should be noted that $e^m$ is the implicit function that represents the shape of the $m^{th}$ obstacle, which is already given by \eqref{eq:ellipse}. When the ego vehicle is in collision with the obstacle, the value of $e^m$ becomes zero and hence $P_{i}^m$ equates to $\tau$. To make the obstacle as non-crossable for the vehicle, this information is used to formulate a constraint in MPC formulation.

The potential field approach for collision avoidance ensures that the size of MPC problem stays constant irrespective of the number of obstacle, an advantage over the spatial constraint approach.

\subsection{Model Predictive Control Formulation}
We rely on MPC to achieve the navigation objectives. The MPC controller repeatedly solves a constrained optimization problem to compute an optimal sequence of control commands that minimize a desired cost function $J$ over a finite horizon $N$. At every sampling instance, only the first element of this control sequence is applied in closed loop. This section details our MPC problem formulation. 

The objectives of the controller are \emph{(i)} to match the human operator's inputs (i.e., $\delta$ and $v$), \emph{(ii)} ensure collision avoidance, and \emph{(iii)} restrict  the controller authority to intervene to avoid surprises for the human operator. To achieve the objectives above we define the following MPC:
\begin{subequations}
\begin{align}
\underset{z, u, S}{\min}  &
    \sum_{k=1}^N J_k(z_k,u_k, S_k)  \label{eq:mpc_a}
\end{align}
\begin{align}
\text{subject to} \quad & z_{k+1} = f(z_k, u_k) \label{eq:mpc_b}\\
& z_0 = z(t)                                    \label{eq:mpc_d}\\
& \delta_{\textrm{min}}         \leq \delta_{k+1}           \leq \delta_{\textrm{max}}      \label{eq:mpc_e}\\
& v_{\textrm{min}}              \leq v_{k+1}                \leq v_{\textrm{max}}           \label{eq:mpc_f}\\
& \Dot{\delta}_{\textrm{min}}   \leq \dot{\delta}_k     \leq \Dot{\delta}_{\textrm{max}}\label{eq:mpc_g}\\
& a_{\textrm {min}}       \leq a_k          \leq a_{\textrm{max}}                       \label{eq:mpc_h}\\
& \resizebox{0.65\hsize}{!}{$
\delta_{\textrm{dev min}} - s^{\delta}_{k+1}     \leq \delta_{k+1} - \delta_{\textrm{ref}}(t) \leq \delta_{\textrm{dev max}}  + s^{\delta}_{k+1}
$}\label{eq:mpc_i}\\
& P_{k+1,i} \leq \tau + s^{p}_{k+1}  \label{eq:mpc_j}\\
& k\in (0,N-1) \notag, 
\end{align}
\end{subequations}
where $z_k$,  $u_k$, and $S_k$ are the state, command, and slack variables (discussed below) at prediction step $k$. The MPC controller minimizes the cost \eqref{eq:mpc_a} subject to the constraints~\eqref{eq:mpc_b}-\eqref{eq:mpc_j}. The cost $J_k$ is defined as follow:
\begin{equation}
   J_k:=  J^{P}_k + J^{\delta}_k +J^{v}_k+ J^{S}_k.
\end{equation}
The first term of the cost function, namely $J^{P}$, is associated with the potential field to ensure a collision free motion and it is defined as follows: 
\begin{equation}
    J^{P}_k = W_{P} P_{k}(x_k,y_k),
\end{equation}
where $W_P$ is a weighting factor and $P$ is defined according to~\eqref{eq:potential_function_all}, for all the circles representing the ego vehicles and for all the obstacles. 

\noindent The second term of the cost function penalizes the deviation of controller input $\delta$ from the reference or human operator's input $\delta_{\textrm{ref}}$ with the penalty $W_{\delta}$,
\begin{equation}
    J^{\delta}_k = W_{\delta} (\delta_{\textrm{ref}}(t) - \delta_{k})^2
\end{equation}
The third term of the cost function penalizes the deviation of controller input $v$ from the reference or human operator's input $v_{\textrm{ref}}$ with the penalty $W_{v}$
\begin{equation}
    J^{v}_k = W_{v} (v_{\textrm{ref}}(t) - v_{k})^2
\end{equation}
The last term in the cost function is used to penalize, with a penalty $W_{s}$, slack variables $S_k:=[s^{{\delta}}_{k},\,s^{p}_{k}]^{\textrm{T}}$ which are used to soften constraints~\eqref{eq:mpc_i} and~\eqref{eq:mpc_j}:
\begin{equation}
    J^{S}_k = W_{s} S_{k}^2.
\end{equation}
Constraint \eqref{eq:mpc_b} represents the dynamic coupling. In addition, note that $f(z_k, u_k)$ represents the discritized kinematic bicycle model equations presented in equations \eqref{eq:kinematic}. 
Constraints \eqref{eq:mpc_e} and \eqref{eq:mpc_g} limits the RWA $\delta$ and its rate of change $\dot \delta$, respectively. Furthermore, Constraints \eqref{eq:mpc_f} and \eqref{eq:mpc_h} limit the velocity $v$ and acceleration $a$ of the vehicle, respectively.  
Constraint \eqref{eq:mpc_i} restrict the controller's authority to deviate from the reference RWA at time $t$. This makes the controller action moderate in the lateral sense and helps avoid larger mismatch between human operator's intentions and the controller's action.
Finally, together with the potential field in the cost function, constraint \eqref{eq:mpc_j} helps ensure collision avoidance by making the obstacles uncrossable. The upper limit of this constraint is governed by $\tau$ as found in \eqref{eq:pot_function}. Constraint \eqref{eq:mpc_j} is softened to practically avoid infeasible solutions, while a high penalty on the slack variables in the cost function limits such a constraint violation.

\begin{figure}[t]
    \centering
    \includegraphics[scale= 0.21]{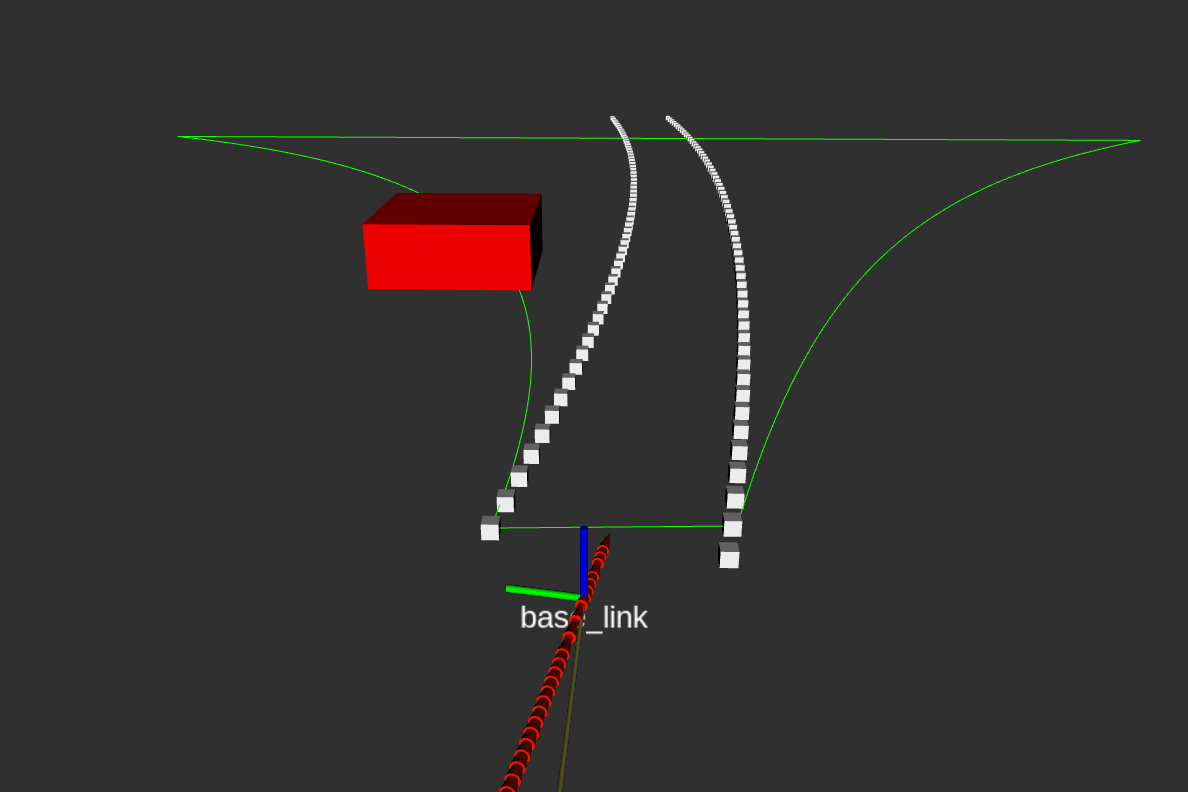}
    \caption{Rviz\protect\footnotemark illustration of visual feedback to be added on top of video stream on the screens at workstation. The maximum and minimum vehicle trajectory due to the corresponding controller intervention is shown by the green
    controller authority cone. Obstacles are represented by red cubes.}
    \label{fig:ros}
\end{figure}
\footnotetext{http://wiki.ros.org/rviz}

\section{MPC DATA-BASED VISUAL FEEDBACK}\label{sec: section3}
To further raise human operator's trust and understanding of the system behaviour, a feedback can be used to communicate the intention of system to the human operator as discussed in \ref{sec:RelatedWork}. Our proposed visual feedback contains information on the region of future vehicle positions due to controller authority, as defined by constraint \eqref{eq:mpc_i}, and future vehicle path, shown by white tracks, due to complete sequence of steering wheel angle and velocity inputs by controller. This feedback is proposed to be added as graphical illustration on top of the video stream on screens at workstation. The visual feedback is expected to help the human operator better understand the possible positions of the vehicle according to both the operator's and controller's inputs. It aims at avoiding any surprises or conflict if the controller intervenes. This concept is illustrated in Fig. \ref{fig:ros}.

\section{MPC DATA-BASED PREDICTIVE DISPLAY}\label{sec:predictive}
\begin{figure}[t]
    \centering
    \includegraphics[width=\linewidth]{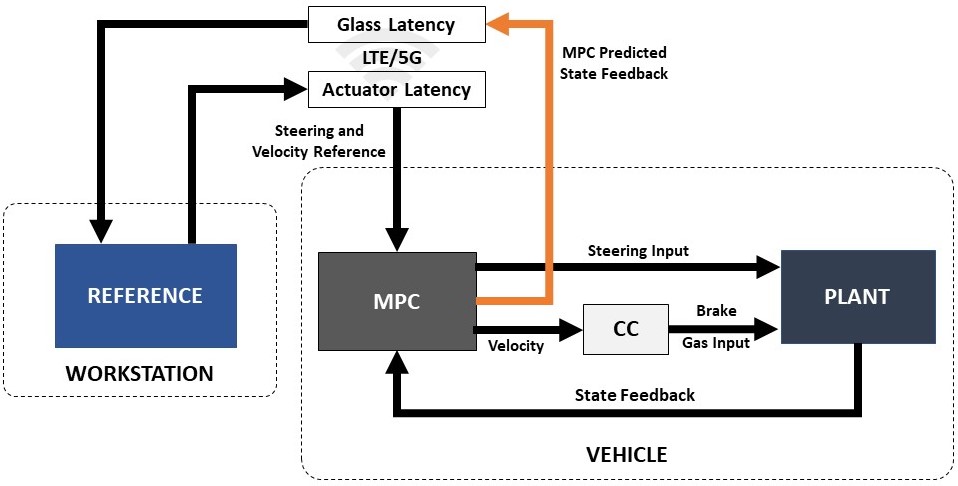}
    \caption{System Architecture for MPC Data-Based Predictive Display.}
    \label{fig:pred arch}
\end{figure}
This section discusses our proposed predictive display method. Our predictive display exploits the MPC predictions to mitigate the effects a high network latency can have on the teleoperation. Recall that the reference module in Fig. \ref{fig:general architecture} generates reference inputs based on the state feedback it receives, which is delayed by time equal to glass latency. These reference inputs are further delayed by a time equal to the actuator latency during the transmission to the MPC module. During this time the ego vehicle would have travelled by time equal to this total round trip latency. The goal of our predictive display is to attempt predicting the ego vehicle position during the time frame of a total round trip latency. This predicted vehicle position can be used to enrich the camera images before they are displayed on the screens at the workstation. 

Authors in \cite{ftmlatency} propose a \emph{full prediction} method, which relies on a bicycle model (as presented in \eqref{eq:kinematic}) for prediction assuming constant vehicle velocity and RWA. This method is taken as baseline for comparison with the proposed MPC data based Predictive Display in section \ref{sec:predictive results}.

In our approach, the idea is that MPC's predicted state corresponding to time equal to total round trip latency could be used as state feedback.
This simplifies the process and needs no new calculations as in \cite{ftmlatency}. MPC already uses a model of the system and sets of constraints and objective to calculate optimal evolution of states and inputs. This method also eliminates the assumptions of constant velocity and RWA. Fig. \ref{fig:pred arch} illustrates this concept.

\section{SIMULATION RESULTS}\label{sec: section5}
The validation of the approach was carried out with simulations in MATLAB/Simulink\textsuperscript{\textregistered}. For simulations, a high fidelity car model from IPG CarMaker~\cite{cm} was used as plant to test the designed controller. A Volvo XC90 T6 AWD was used as the testing vehicle. The optimization of MPC problem is carried out through \texttt{acados}, an open source software package with a collection of solvers for fast embedded optimization intended for fast embedded applications \cite{acados}. Sequential Quadratic Programming (SQP) is used as the solver by \texttt{acados}. The algorithm runs well below \SI{50}{ms} on an Intel Core i7-4600 2.10 Ghz 2 Core CPU with 8 GB of RAM and hence it can be expected to perform even better when run on a powerful hardware of the experimental vehicle. The system architecture for ToD is shown in Fig. \ref{fig:general architecture}. The architecture consists of two main blocks, that are, \emph{(i)} the workstation, where the human operator generates reference inputs of velocity and RWA, \emph{(ii)} the teleoperated vehicle equipped with the proposed MPC controller. In a real setup, the signals between these two blocks are transmitted over a commercial 4G LTE network and hence delayed due to network latency.
In a simulation setup, delay blocks are added separately to artificially introduce this latency in simulations.
For simulations, the reference block generates reference velocity, whose values are predefined as a function of time. In addition to this, the reference block simulates the human operator using a Feedback Linearized Path Tracking Controller (FBLC), taken from \cite{fblc1}. The control law used to simulate the human operator is defined as follows
\begin{equation}
    \delta_{\textrm{FBL}}(t)  = \arctan\Big(\frac{-\gamma_{1} e_{L}(t) - \gamma_{2} v \sin{(e_{H}(t))}}{  v^2 \cos(e_{H}(t))}\Big)
\end{equation}
where, the lateral and heading errors are $e_{L}$ and $e_{H}$, respectively, and $v$ is the current vehicle velocity. The correction for heading error and lateral error can be prioritized using the gains $\gamma_{1}$ and $\gamma_{2}$. The errors, $e_{L}$ and $e_{H}$ are computed with respect to the reference tracking point along the trajectory, which is again a function of lookahead distance. To include the effect of visual feedback to the human operator, we added another feedback term. This term penalizes the deviation between the operator SWA (RWA) or the calculated $\delta_{\textrm{FBL}}$ and the current SWA (RWA), with the proportional gain $\gamma_{3}$. Thus, the overall simulated human operator's RWA input becomes:
\begin{equation}
    \delta_{\textrm{ref}}(t)  =  \delta_{\textrm{FBL}}(t) + \gamma_3(\delta(t_{0}) - \delta_{\textrm{FBL}}(t))
\end{equation}

The CarMaker plant cannot take velocity input directly and hence a Cruise Control (CC) is designed to translate the desired velocity from MPC into corresponding Gas and Brake pedal values. A basic CC is formulated based on a proportional-integral controller. 

The sampling time of MPC is set to $t_s=$ \SI{50}{ms} since the solution times are well within this limit. The order of the ellipse $n$ is taken as \si{4}. The potential function parameters are tuned to $\tau=$ \num{0.1} and $\rho=$ \num{2}. This tuning results in a close manoeuvre of the vehicle around the obstacles. 
ToD is expected to be performed at low speed due to the perils of remote driving at high speed. Therefore, these simulations consider only low speed (\SI{3}{m/s}) scenarios where the effect of latency is not so crucial and hence a constant latency in the simulation is considered. In the following, the performance of the proposed controller, namely, our Active Safety System (ASS), is compared with a baseline approach which does not have longitudinal control capability and restriction on deviation of RWA from its corresponding reference. Section \ref{sec: overtake} compares the two approaches in a scenario that involves overtaking a stationary obstacle with another dynamic obstacle in the scene. In addition, Section \ref{sec:predictive results} illustrates our MPC-data-based predictive display concept.

\subsection{Overtake Scenario}\label{sec: overtake}
\begin{figure}[]
    \centering
    \includegraphics[width=\linewidth]{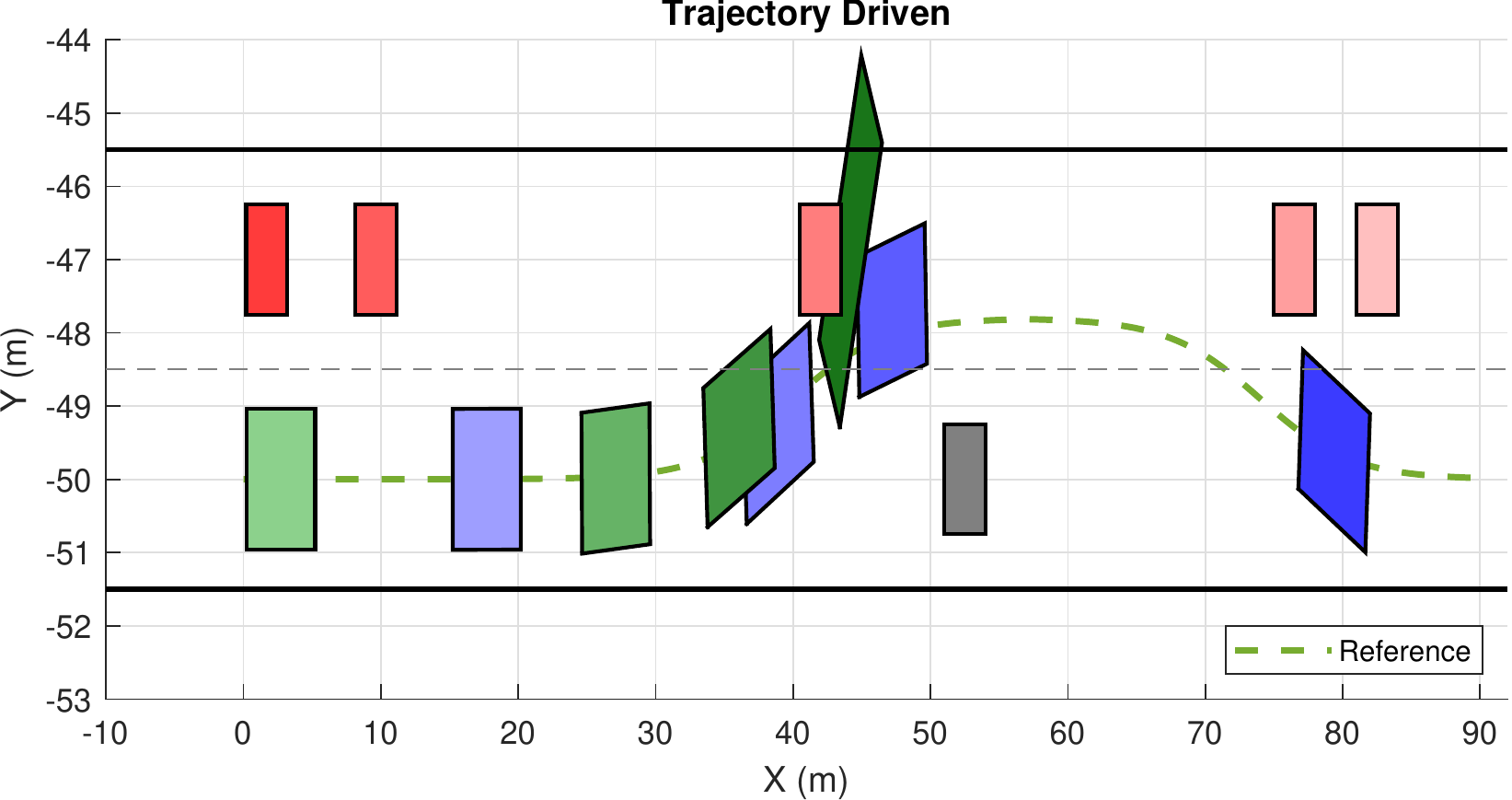}
    \caption{Trajectory Driven for Overtake Scenario. Progression of ego vehicle with ASS is shown in shades of blue. Progression of ego vehicle with Baseline controller is shown in shades of green. Progression of oncoming obstacle (dynamic) is shown in shades of red. Stationary obstacle is shown in grey color. The shades go from light to dark with the progression of time.}
    \label{fig:overtake fig1}
\end{figure}
\begin{figure}[t]
    \captionsetup{justification=centering}
     \centering
     \begin{subfigure}[b]{\linewidth}
         \centering
        \includegraphics[width=\linewidth]{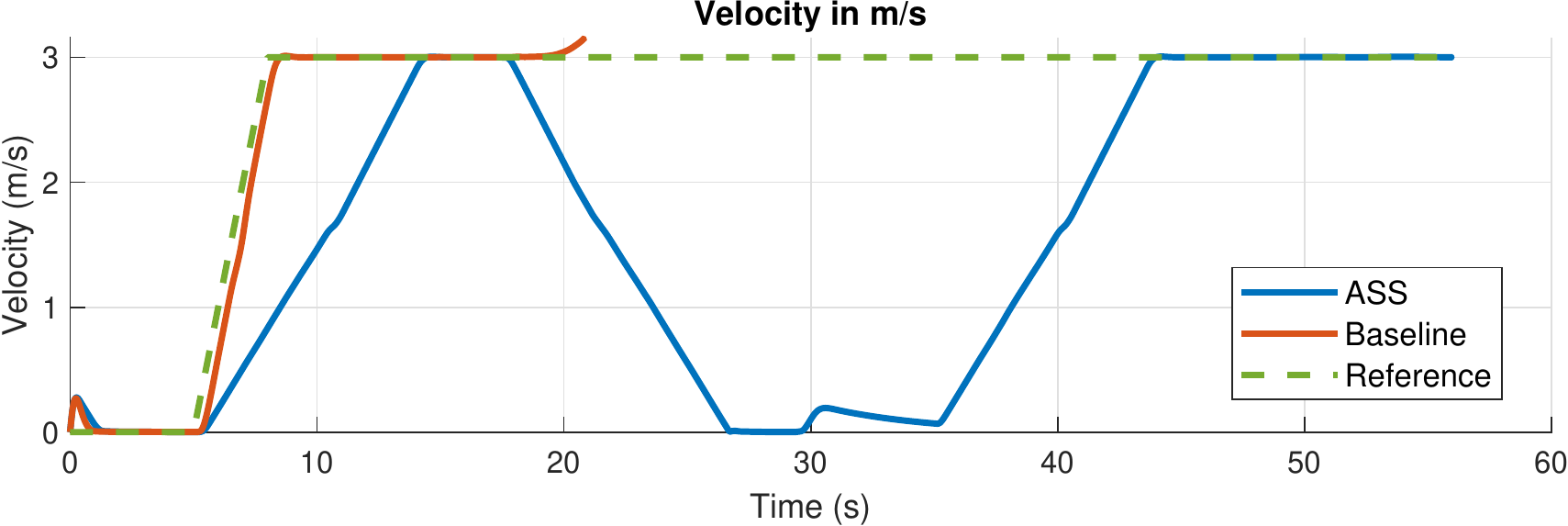}
        \caption{Velocity}
        \label{fig:overtake v}
     \end{subfigure}
     \newline
     \begin{subfigure}[b]{\linewidth}
         \centering
         \includegraphics[width=\linewidth]{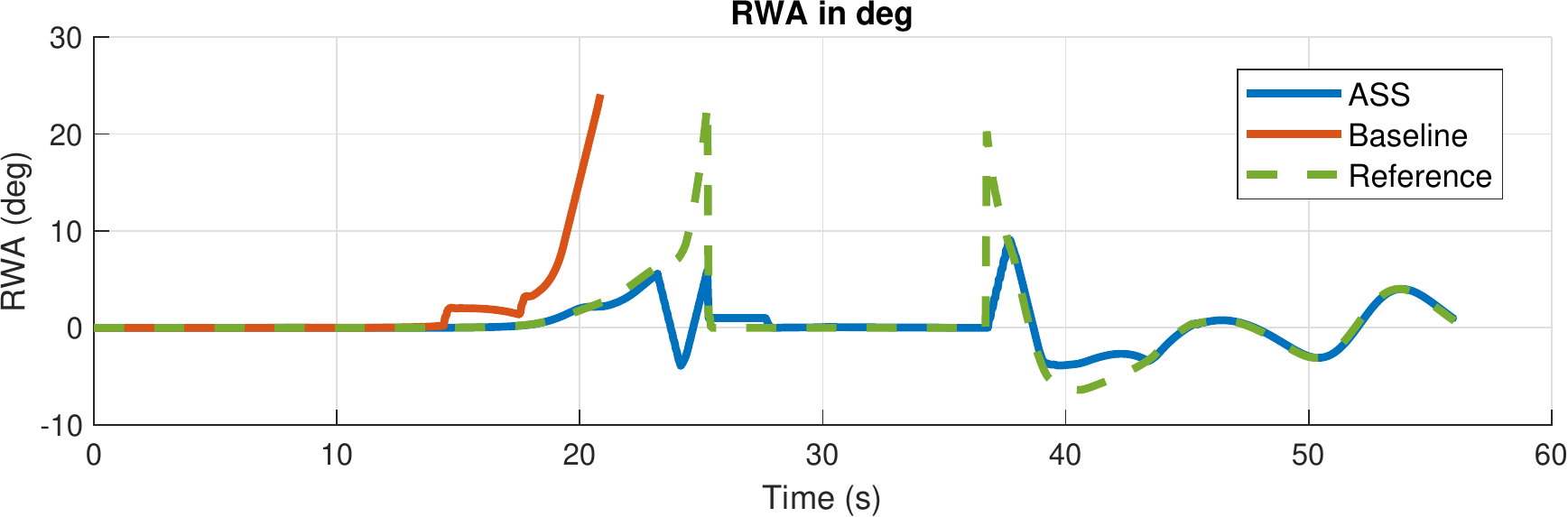}
        \caption{RWA $\delta$}
        \label{fig:overtake delta}
     \end{subfigure}
        \caption{Longitudinal velocity and RWA of the ego vehicle in the overtake scenario.}
        \label{fig:overtake v and delta}
\end{figure}

This scenario considers the ego vehicle trying to overtake a stationary obstacle along its path (e.g., a parked car). While overtaking, the ego vehicle has to deal with an upcoming vehicle proceeding in the opposite lane.
At every sampling time, the dynamic obstacles are elongated in size over the prediction horizon assuming their current heading and velocity stays constant. Effectively converting them into large static obstacles with changing locations and dimensions. Thus, collision avoidance of the dynamic obstacles is accounted for in a conservative manner, with a possibility of reducing this conservatism by accounting for predicted obstacle positions along the horizon by a perception module.
Fig. \ref{fig:overtake fig1} shows the behavior of the ego vehicle with our ASS approach. The reference path (green dashed line) of the ego vehicle's CoM is purposefully designed in such a way that the ego vehicle collides with the stationary obstacle while going around. The ASS-based ego vehicle brakes when encountering the oncoming obstacle, and then overtakes the stationary obstacle when there is free space. This behavior is reflected in the velocity and RWA results in Fig.~\ref{fig:overtake v and delta}, which compares our method (blue lines) with the baseline controller (red lines). The velocity plot in Fig. \ref{fig:overtake v} highlights the contribution of our ASS on the control of the vehicle. As the figure shows, our system reduces the vehicle velocity to prevent collisions, in contrast with the constant reference signal provided by the operator. As Fig. \ref{fig:overtake delta} highlights, our ASS avoids collision by correcting the RWA at time \SI{40}{s}, where the RWA differs from reference RWA. Thus, longitudinal and lateral control capabilities of the ASS are highlighted here. The vehicle with Baseline controller fails to stop after encountering the oncoming obstacle and collides with it since it does not have any longitudinal control capability. The end of its trajectory represents the point where the vehicle leaves the road boundary and hence the simulation gets aborted by CarMaker. This highlights the drawback of not having longitudinal control capability. In addition to this, the RWA plot in Fig. \ref{fig:overtake delta} also shows a large deviation of RWA from reference RWA. This steering behaviour is highly counter-intuitive to what the human operator wishes to do, potentially raising trust issues over the system. In comparison to this, even though the ASS steering profile is showing considerable deviation from the reference RWA, it remains well around the range of reference and follows the reference back again. This highlights the importance of restricting the controller's authority to deviate from the reference RWA by using the constraint \eqref{eq:mpc_i}.

\subsection{Predictive Display}\label{sec:predictive results}
\begin{figure}
    \centering
    \includegraphics[width=\linewidth]{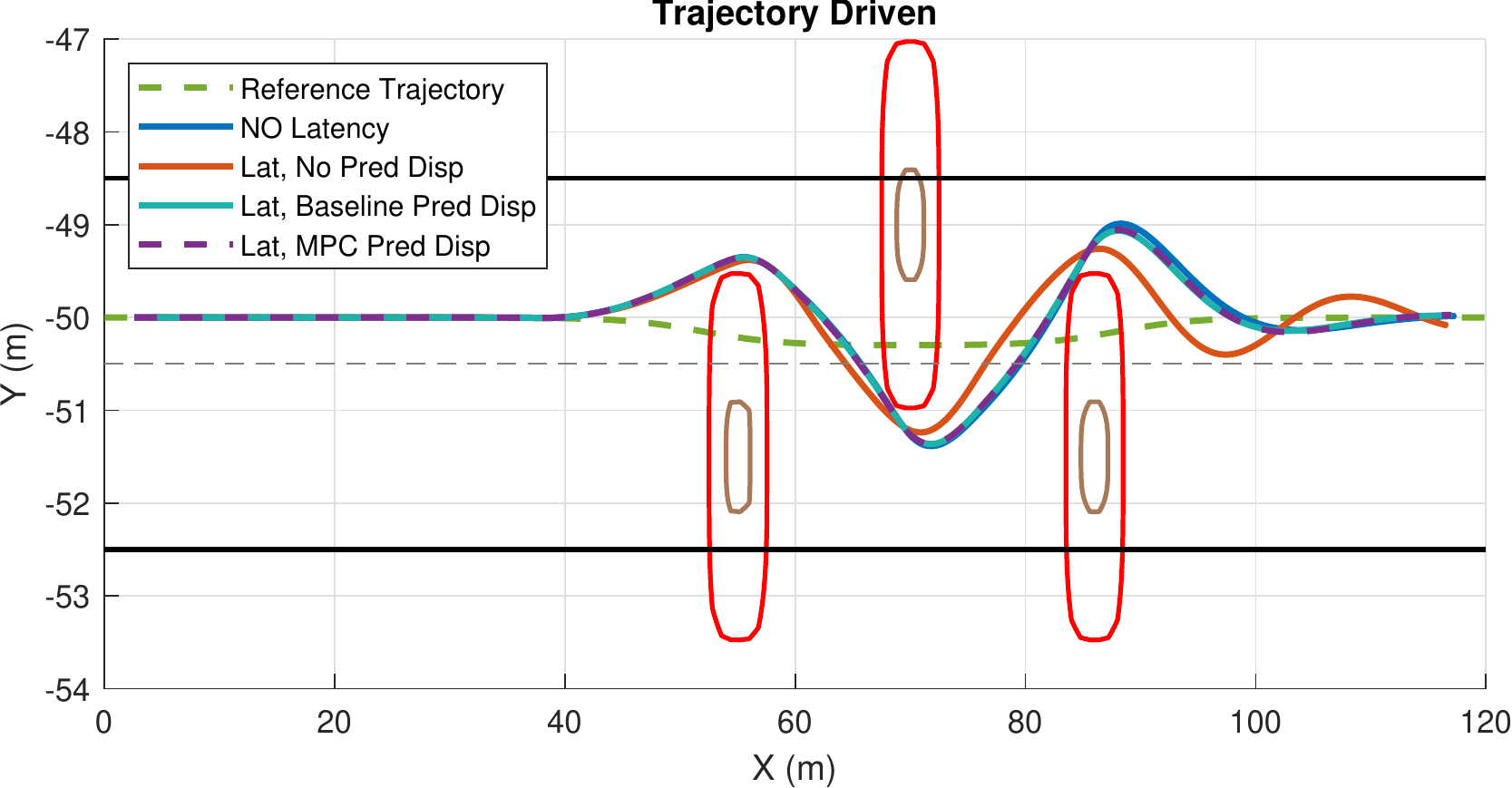}
    \caption{Trajectory of the CoM of the ego vehicle driving through closely placed obstacles with latency.}
    \label{fig:predictive fig1}
\end{figure}

The predictive display is expected to nullify the effect of latency and produce results equivalent to that for no latency in system. Fig. \ref{fig:predictive fig1} compares the performance of our ASS with our proposed predictive display (dashed purple line) with the baseline approach~ \cite{ftmlatency} (solid light blue line) and our ASS without predictive display (solid red line). The performance of our method with no latency is used as a benchmark. In addition, the  stationary obstacles are represented by higher order ellipse (brown) and they are inflated by the radius of the ego circle as represented by the large ellipses (red). The reference trajectory (green dashed line) is purposefully designed for a collision with obstacles. In this simulation, we consider a total round trip latency of \SI{500}{ms}.
The trajectory driven by the CoM of the ego vehicle for both the cases with predictive displays (Baseline and MPC data-based) overlap the trajectory with no latency in the system. This proves that both predictive display techniques equally nullify the effect of latency. Nevertheless, the advantage of the proposed MPC-data-based predictive display method over the baseline method lies in the fact that it does not assume constant velocity and RWA for predicting vehicle position. It also removes extra calculation step of predicting vehicle position using a bicycle model as done in \cite{ftmlatency}. 

It should be noted that, even in the absence of any latency mitigation technique, the vehicle trajectory stays collision free, verifying that the controller is efficient to avoid collision even with delayed (outdated) reference inputs.

\section{CONCLUSION}
This paper presented an ASS approach for teleoperated vehicles based on model predictive control. The ASS controls both the lateral and longitudinal vehicle's dynamics allowing the vehicle to perform a large range of maneuvers and avoid collision by both braking and steering intervention. Collision avoidance is achieved by using artificial potential fields in the MPC controller. Our ASS also has an added restriction on the controller's authority to prevent large deviations from the human operator's RWA inputs. In addition, we proposed a visual feedback in the form of graphical presentation on the monitors of workstation. 
Finally, to mitigate the effects of large latency in a teleoperation setup, we proposed a novel MPC data-based predictive display method. We tested our approach by using a high fidelity vehicle model for plant and a realistic environment.
The simulation results highlighted the advantages of our ASS for teleoperation compared to existing baselines. Future work includes experimental validation of the proposed work.




\newpage
\section*{APPENDIX}

The MPC penalties and FBLC gains for the corresponding scenarios are summarized in the following table. 

\begin{table}[h!]
\centering
\caption{MPC penalties and FBLC gains.}
\begin{tabular}{|l|c|c|c|c|c|c|c|}
\hline
                   & \multicolumn{1}{l|}{$W_{P}$} & \multicolumn{1}{l|}{$W_{\delta}$} & \multicolumn{1}{l|}{$W_{v}$} & \multicolumn{1}{l|}{$W_{s}$} & \multicolumn{1}{l|}{$\gamma_1$} & \multicolumn{1}{l|}{$\gamma_2$} & $\gamma_3$ \\ \hline
Overtake Scenario  & 0.1                          & $10^3$                            & 1                            & $10^5$                       & 1                               & 2                               & 0.25       \\ \hline
Predictive Display & 0.1                          & $10^2$                            & 1                            & $10^5$                       & 1                            & 2                             & 0.25       \\ \hline
\end{tabular}
\end{table}


\section*{ACKNOWLEDGMENT}
This work was presented at the workshop for Road Vehicle Teleoperation (WS09), IV2021.

\bibliographystyle{IEEEtran}
\bibliography{bibliography}

\end{document}